# Sentiment Informed Sentence BERT-Ensemble Algorithm for Depression Detection

Bayode Ogunleye [1,*], Hemlata Sharma [2] and Olamilekan Shobayo [2,*]

1. Department of Computing & Mathematics, University of Brighton, Brighton BN2 4GJ, UK
2. Department of Computing, Sheffield Hallam University, Sheffield S1 2NU, UK; h.sharma@shu.ac.uk
* Correspondence: b.ogunleye@brighton.ac.uk (B.O.); o.shobayo@shu.ac.uk (O.S.)

**Abstract:** The World Health Organisation (WHO) revealed approximately 280 million people in the world suffer from depression. Yet, existing studies on early-stage depression detection using machine learning (ML) techniques are limited. Prior studies have applied a single stand-alone algorithm, which is unable to deal with data complexities, prone to overfitting, and limited in generalization. To this end, our paper examined the performance of several ML algorithms for early-stage depression detection using two benchmark social media datasets (D1 and D2). More specifically, we incorporated sentiment indicators to improve our model performance. Our experimental results showed that sentence bidirectional encoder representations from transformers (SBERT) numerical vectors fitted into the stacking ensemble model achieved comparable F1 scores of 69% in the dataset (D1) and 76% in the dataset (D2). Our findings suggest that utilizing sentiment indicators as an additional feature for depression detection yields an improved model performance, and thus, we recommend the development of a depressive term corpus for future work.

## 1. Introduction

There is a growing concern as regards the deteriorating mental health of people [1,2]. WHO [3] reported a growing drop in the mental health of people as individuals diagnosed with depression has increased. In recent times, depression has garnered global attention due to its widespread prevalence and the significant risk it poses, particularly in terms of suicide [1,3,4]. Depression is one of the leading causes of disability globally [5]. WHO [3] reported approximately 280 million people in the world suffer from depression. Their findings showed more women are affected than men, and 5% of adults suffer from depression. The literature evidences the symptoms of depression as unhappiness, loneliness, forgetfulness, low self-esteem, thoughts of self-harm, disrupted sleep, loss of energy, changes in appetite, anxiety, reduced concentration, indecision, feelings of worthlessness, lack of interest, guilt, or hopelessness [2,6]. Previous studies showed that the primary causes of depression are financial issues, workplace issues, family issues, and academic performance [7]. There are traditional approaches to depression detection that require psychometric questions. Several depression questionnaires have been developed. For example, Beck's Depression Inventory [8], Center for Epidemiologic Studies Depression Scale [9], Children's Depression Inventory [10], Mood and Feelings Questionnaire [11], Patient Health Questionnaire-9 [12], and Revised Children's Anxiety and Depression Scale [13]. However, this approach is labor-intensive, time-consuming, and limited to pre-defined attributes. In addition, research evidenced that people with depression may not know or acknowledge they have the illness and thus influences the survey outcome for early-stage depression screening [14,15]. The ability to detect depression in its early stages is important in providing prevention, diagnosis, or treatment. Corrective treatment is usually too late as the sufferer's quality of life might have been affected adversely [1]. This can be attributed to the sufferer's attitude to seek medical help at the early stages of the disease, making their symptom worsen. Liu et al. [16] stated that 70% of sufferers will not seek medical help for their depressive conditions. Social media, however, has been a proven source of obtaining public opinion of people, providing insight into the conditions of their mental wellness [17]. People obtain information or express their feelings freely on social media [1,18,19]. Thus, user-generated content (UGC) is considered a reliable data source for developing automated depression detection models [20,21]. Interestingly, the rapid development of machine learning (ML) approaches has made significant contributions to the development of depression detection algorithms [22]. Thus, this study will focus on the use of ML techniques.

Several ML algorithms were implemented for depression detection, and so far, traditional algorithms and predominantly only one stand-alone algorithm have been used [23–32], which are unable to deal with data complexities, prone to overfitting, and limited in generalization, and thus, the effectiveness of the models is unsatisfactory for real-world deployment [1]. Most importantly, a major concern of existing algorithms is the generalization issue [4,21,33–35]. Based on this background, our aim is to develop a well-performing ML algorithm for depression detection. To this end, we propose the use of the Sentence BERT-Ensemble model for depression detection. Thus, our main contributions can be summarised as follows.

1. We conduct an experimental comparison of several state-of-the-art (SOTA) ML algorithms for depression detection and discuss them from a scientific lens.
2. We demonstrate the use of the Sentence BERT-Ensemble model to achieve SOTA results.
3. We demonstrate that the sentiment analysis indicator is a useful external feature in depression detection.

Subsequent sections present the background knowledge of this study (Section 2), methodology (Section 3), and results (Section 4), and Section 5 will provide conclusions and recommendations.

## 2. Related Works

Previous studies have proposed several statistical and ML methods for depression detection. For example, K-nearest neighbors [23], support vector machine [24], decision tree [25], XGBoost [26], multinomial Naïve Bayes [27], convolutional neural network [28], recurrent neural network [29], long short-term memory [30], Bidirectional long short-term memory [31] and RoBERTa [32]. However, the problem of generalization in existing depression detection algorithms [4,21,33–35] has been a major concern. For illustration, Cai et al. [4] collected data from the Chinese social network platform Sina Weibo version 14.9.0. The dataset consists of 742,430 depressed tweets and 729,409 non-depressed selected from 3711 depressed users and 19,526 non-depressed users. Thus, the limitation of their study is a generalization issue. This is because depressed users tend to post fewer tweets than non-depressed users on Twitter, while users on Sina Weibo behave just the opposite.

The use of traditional classification methods may not succeed due to the difficulty in extracting discriminative features from texts on social media [1,35]. Thus, studies like Shrestha et al. [36] applied five variants of bidirectional encoder representations from transformers (BERT) with and without fine-tuned layers for depression detection. Their study focused on identifying moderate depression using typed and transcribed responses from the StudentSADD dataset. Their findings highlighted that typed responses hold more value than transcribed responses in depression screening. They showed DistilBERT with fine-tuning layers demonstrated the best performance for typed responses with an accuracy of 63%. Naseem et al. [37] formulated depression detection as a multi-class classification task by recategorizing the depression severity levels as minimal, mild, moderate, and severe. They showed embeddings from text graph convolutional network (TextGCN) feed into Bidirectional long short-term memory (BiLSTM) with attention layer, and ordinal classification layer outperforms support vector machine (SVM), random forest (RF), multilayer perceptron (MLP), CNN, recurrent neural network (RNN), long short-term memory (LSTM), BiLSTM and DepressionNet across two different Reddit datasets. Their approach achieved an F1 score of 85% (in their D1 dataset of 3553 posts) and 95% (in their D2 dataset of 77,350 posts). Pérez et al. [21] made publicly available BDI-Sen dataset comprising 4973 annotated sentences covering depressive symptoms and 41,200 control sentences for depression detection research. In their study, they showed that MentalBERT (MBERT) outperformed BERT, BERT-mini, T5 (text-to-text transfer transformer), logistic regression (LR), and SVM as a binary classification task with an f1 score of 83%. Similarly, they performed a fine-grained depression severity level and showed MBERT outperformed other models. Monreale et al. [38] used the psychometric profile of users as a feature to improve the performance of SVM for depression detection and, thus, achieved an accuracy of 96%. Sen et al. [39] aimed to understand the effect of company and region on measuring workplace depression. Their study adopted the occupational depression inventory (ODI) scale designed for work-related depression assessment. They collected over 358,527 employee reviews from 104 prominent US companies from 2008 to 2020. In addition, the yearly stock growth values of the 104 companies were collected from the Yahoo financial portal as values of five social economic indicators. Thus, the proposed AutoODI1 framework which uses SBERT and cosine similarity to assign composite ODI scores to the reviews. Their findings suggest US states that host companies with high ODI scores also manifested high depression rates, talent shortage, and economic deprivation. Wu et al. [40] proposed a deep neural network for the early prediction of depression risk. Their model leveraged daily mood fluctuations as a psychiatric indicator and incorporated textual and emotional attributes through knowledge distillation and achieved an AUROC of 0.9317 and AUPRC of 0.8116. Figuerêdo et al. [1] proposed a convolutional neural network (CNN) in combination with context-independent word embeddings and fusion-based approaches. They used a dataset collected from Reddit users (in English). In their experiment, they trained the algorithms using 30,851 depressed posts and 264,172 non-depressed. Thereafter, they tested with 18,706 depressed posts and 217,665 non-depressed, and their approach achieved a precision of 76% and an F1 score of 71%.

Moreover, Villatoro-Tello et al. [41] prepared a lexicon and used it as a feature for the development of a depression detection binary classification model. They used the patient's health questionnaire (PHQ-8) score to determine the level of the patient's depression. They performed their experiment using the Distress Analysis Interview Corpus—wizard of Oz (DAIC-WOZ) dataset and the Extended Distress Analysis Interview Corpus (E-DAIC), which is an extended version of the DAIC-WOZ dataset and achieved a macro F1-score of 83% and 80%, respectively. They showed their lexical availability theory approach outperformed BERT and MLP in a clinical interview context. However, it is worth stating that their implementation did not handle the class imbalance issue, which might impact generalization. Zhang et al. [6] proposed a Span-based PHQ-aware and similarity contrastive network (SpanPHQ) for identifying depressive symptoms. Their model took into consideration both semantic contextual features and PHQ-9 descriptive features. Their approach evidenced superior performance compared to BERT and its variants. Liu et al. [42] argued that ecological momentary assessment (EMA) captures the mood better than standard questionnaires like PHQ-9. To evidence this, they analyzed the EMA data and passive sensing data to generate synergy, which can be deployed to enhance the performance of depression mood detection. They used their approach to enhance several ML algorithms, including SVM, LR, KNN, RF, and decision tree (DT). Their models were applied to the StudentLife dataset and achieved an average accuracy of 98%. Malik et al. [43] used the Hamilton rating scale to determine the level of people's depression; the survey results (1694 records) were modeled for depression detection using ML algorithms, including KNN, DT, and naïve Bayes (NB). They showed that KNN was superior, with an accuracy of 92%. Authors Gallegos-Salazar et al. [44] proposed a contrast pattern-based classifier to detect depression by using a new data representation based only on emotion and sentiment analysis. Similarly, Amanat et al. [30] showed that RNN-LSTM outperformed SVM, NB, CNN, and decision trees with an accuracy of 99%.

Burdisso et al. [45] proposed an SS3 (smoothness, significance, and sanction) framework to improve incremental classification, support for early classification, and explainability for the text classification algorithm. They compared SS3 to LR, SVM, KNN, and NB. Their approach was achieved with an F1 score of 61%. Trotzek et al. [46] proposed a hybrid approach for depression detection. In their study, they developed an LR classifier by incorporating readability and emotion features into user-level linguistic metadata and enhanced its performance using CNN.

Adarsh et al. [47] addressed the problem of imbalance present across various age groups and demographics through the utilization of the one-shot decision approach. They used an innovative ensemble model that combines SVM and KNN techniques and achieved an accuracy of 98.05%. Guo et al. [48] constructed the Chinese depression lexicon by expanding the Dalian University of Technology sentiment lexicon (DUT-SL) with depression-specific vocabulary (with the help of domain expertise). They proposed an ensemble method based on the newly constructed lexicon and feature fusion approach. Thus, their approach outperformed ML algorithms such as logistic regression, RF, KNN, SVM, DT, eXtreme Gradient Boosting (XGBoost), and light gradient boost machine (LightGBM). To conclude, our review of the literature findings showed that previous studies have generalization problems, and mostly one stand-alone algorithm has been developed for depression detection algorithm. These approaches struggle with data complexities, are prone to overfitting, and are limited in generalization; thus, the effectiveness of the models is unsatisfactory for real-world deployment [1]. Furthermore, the literature synthesis showed that few studies employed ensemble methods. It is worth stating that the unique characteristic of our work, as opposed to the previous works, is instead the combination of sentiment features and feature representation from sentence BERT fitted into the ensemble learning model. In the next section, this paper details the methodology adopted in this study.

## 3. Methodology

This section presents the methods applied in this study. This study conducted an extensive and thorough comparison of ML algorithms ranging from traditional ML, deep learning, and transformer-based algorithms. Based on the target variable of the datasets used (discussed in Section 3.2), this study formulates the depression detection task as a multiclass classification problem. In our attempt to improve generalization and ascertain the SOTA approach, we performed several experiments using two datasets, as outlined below.

1. In the first experiment, we compared traditional ML algorithms using term frequency and inverse document frequency (TF-IDF) vectorizer.
2. In the second attempt, we compared ML algorithms using contextual word embeddings such as BERT and SBERT.
3. Finally, we implemented sentiment analysis and used the polarity result as an explicit feature. Thus, we compared ML algorithms using contextual word embeddings.

### 3.1. Proposed Approach

This paper compared several ML approaches that have shown good performances in the depression detection literature. Thus, based on our experiment, this study proposes the use of sentence BERT embedding and stacked ensemble model.

The experimental setup of our approach is shown in Figure 1. The raw data are converted into word embeddings in numerical vector models such as BERT. In parallel, sentiment analysis is performed, and thus, the numerical vectors are fed to the classifier (stacked ensemble model) to provide the required prediction. Subsequent sections discuss the individual components of our proposed approach for depression detection, presenting the algorithms used in each stage. For illustration, Sections 3.1.1 and 3.1.2 present the BERT and SBERT for the word embedding component, Sections 3.1.3–3.1.7 present the stacked ensemble model for the classifier component, and Section 3.1.8 presents the Afinn lexicon for the sentiment analysis component.

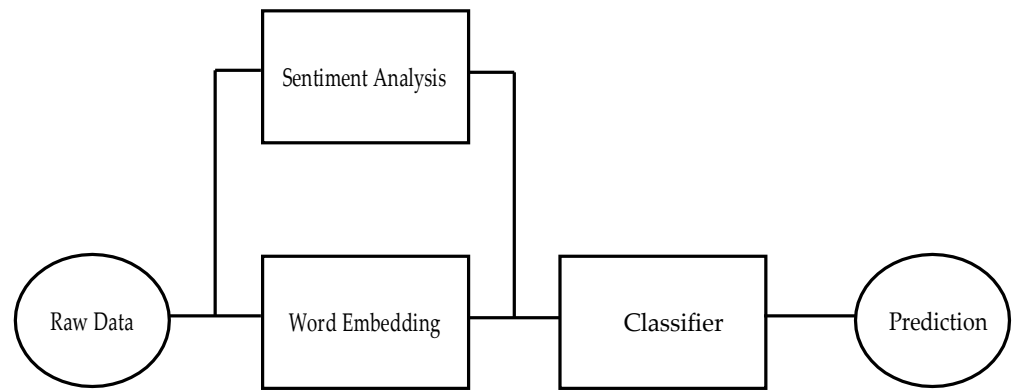

**Figure 1.** An illustration of our experimental setup.

3.1.1. BERT (Bidirectional Encoder Representations from Transformers)

Given a sequence of input embeddings $X = x_1, x_2, \ldots, x_n$ the BERT model [49] consists of an encoder stack that utilizes self-attention mechanisms. The self-attention mechanism involves three sets of weight matrices Query (*Q*), Key (*K*) and Value (*V*). These are used to compute attention scores, and the output of the self–attention layer is obtained by applying these attention scores to the values.

The self–attention mechanism is mathematically represented as:

$$\text{Attention}(Q, K, V) = \text{softmax}\left(\frac{QK^T}{\sqrt{d_k}}\right) V \tag{1}$$

Here, $Q = XW_Q$,
$$K = XW_K$$
$$V = XW_V$$

$d_k$ is the dimensionality of the key vectors.

BERT stacks multiple layers of these self-attention mechanisms to capture conceptual information in the input sequence. BERT uses a masked Language model (MLM) objective during pre-training, where some of the input words are masked, and the model is trained to predict the masked words based on the context.

3.1.2. Sentence-BERT

Sentence BERT (SBERT) [50] is a variant of BERT designed to reduce the complexity of the computational overhead presented in the BERT algorithm for regression task such as sentence-pair similarity. It has also been used to classify sentence pairs based on the labeled class (binary or multi-class). SBERT framework with classification objective (used for this work) consists of a Siamese bi-encoder that comprises of two fine-tuned BERT models, each providing different output vector embeddings *u* and *v*. A third vector $|u - v|$ is also obtained, which is the element-wise absolute difference between the vector embeddings of the Siamese BERT encoder. The vectors are then concatenated and multiplied by a weight matrix *W* and the output is sent to a SoftMax layer to generate the probabilities for each class. This is illustrated in Figure 2.

3.1.3. Stacking Ensemble Model

Ensemble techniques have experienced an increase in relevance in recent years. This can be attributed to the increased performance it provides when compared to using individual models for ML tasks such as regression and classification. It was first introduced in the early 1990s with the idea that individual algorithms with weak predictions could be improved to provide better predictions by combining them in some specific or methodological order [51]. The most common methods used in ensemble learning include Bagging, Boosting, and Stacking. Each of these methods varies slightly with their architecture, and their selection is dependent on the nature of the ML task. For this work, we have used the stacking ensemble framework.

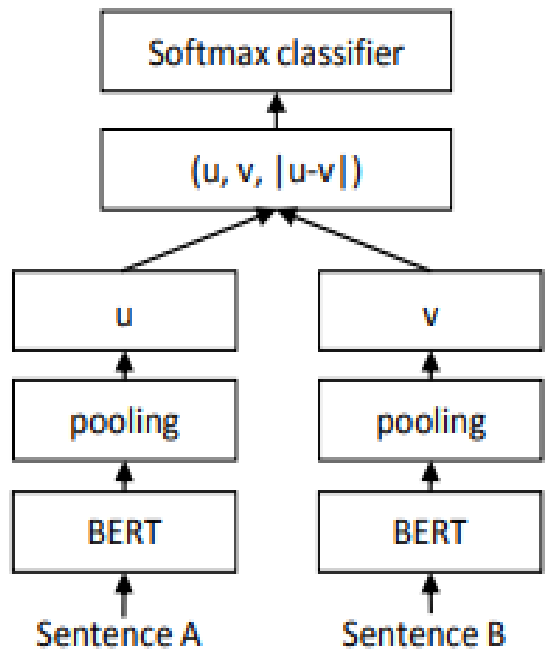

**Figure 2.** Classification objective framework of the SBERT model [50].

Stacking, also known as a stacked generalization, is an ensemble learning strategy that generates predictive outcomes of selected initial learner models, known as the base models, to a final classifier model, known as the meta-classifier, to obtain the final classification for the specific ML task and in most cases, improves the overall performance of the classification when compared to the individual base models and has seen it used in different classification tasks ranging from healthcare to agri-business [52–55]. As illustrated in Figure 3, stacking provides model variety, allowing all base classifiers to participate in the learning process, which in turn helps reduce bias. It also provides interpretability of the learning capacity of the base models and how they influence the results of the selected meta-classifier. The diversity among the chosen base models helps the meta-classifier produce improved classification results. This is due to the different principles of generalization by the base models. The diversity provided by the models is obtained by using learner models of varying learning strategies, thereby introducing a level of disagreement in pattern recognition [56].

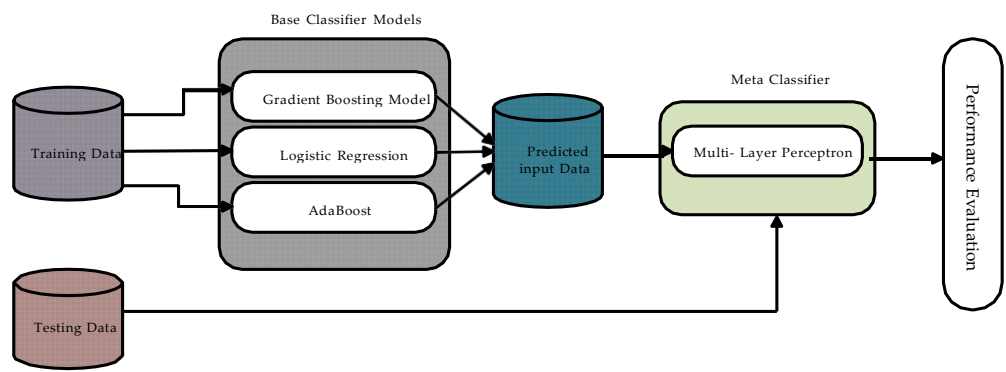

**Figure 3.** Illustration of stacked ensemble model.

The ensemble technique can be represented using the equation:

$$\hat{y} = f_{meta}[b_{mi}(y_i \ldots \ldots \ldots y_N)] \tag{2}$$

where $\hat{y}$ represents the predicted output based on the meta classifier function $f_{meta}$ on the output of the selected individual base models $b_m$. With $y_i$ representing the predicted output of the individual base models. In the subsequent section, we discuss individual ML algorithms deployed.

3.1.4. Gradient Boosting

Gradient Boosting is a powerful ensemble technique in ML. Unlike traditional models that learn from the data independently, boosting combines the predictions of multiple weak learners to create a single, more accurate, strong learner. The key concept is to sequentially add weak learners, each one correcting the errors of its predecessors. Mathematically, the gradient boosting algorithm can be expressed as follows [57]

$$g_t(x) = E_y \left. \frac{\partial \Psi(y, f(x))}{\partial f(x)} \right|_{x} \Bigg|_{f(x) = \hat{f}^{t-1}(x)} \tag{3}$$

where; $\Psi(y, f(x))$ is the loss function measuring the difference between the actual value of $y$ and the predicted value $f(x)$.

$\hat{f}^{t-1}(x)$ is the prediction from the model at iteration $t - 1$.

$g_t(x)$ represents the pseudo-residuals, the negative gradients of the loss function with respect to the current model predictions.

In each iteration, a new weak learner $h^t(x)$ is trained to predict these pseudo-residuals. The model is then updated by adding the new learner, scaled by a learning rate $\eta : f^t(x) = f^{t-1}(x) + \eta h^t(x)$. This process continues for a predefined number of iterations, resulting in a strong predictive model that minimizes the loss function effectively.

3.1.5. AdaBoost

AdaBoost (Adaptive Boosting) is an ensemble learning algorithm that combines the predictions of multiple weak classifiers to create a strong classifier. It works by focusing on the training samples that are hard to classify and adjusting the weights of the weak classifiers accordingly [58].

Given a training set $T = \{T = (x_i, y_i)\}_{i=1}^{N}$ where $x_i$ are the input feature and $y_i$ are the class labels ($y_i \in \{-1, 1\}$), the AdaBoost algorithm can be described as follows:
1. Initialize the Weights:

$$w_i^{(1)} = \frac{1}{N} \text{ for all } i = 1, 2, \ldots, N$$

2. For $t = 1$ to $T$ (number of iterations):
   a. Train a Weak Classifier $h_t$ using the weighted training set.
   b. Compute the Weighted Error $\in_t$:

$$\in_t = \frac{\sum_{i=1}^{N} w_i^{(t)} \times I(h_t(x_i) \neq y_i)}{\sum_{i=1}^{N} w_i^{(t)}} \tag{4}$$

   where $I$ is the indicator function that returns 1 if the condition is true and 0 otherwise.
   c. Compute the Classifier Weight $\alpha_t = \frac{1}{2} \ln \frac{1-\epsilon_t}{\epsilon_t}$
   d. Update the Weights:

$$w_i^{(t+1)} = w_i^{(t)} \exp(-\alpha_t y_i h_t(x_i))$$

Normalize the weights so that they sum to 1:

$$w_i^{(t+1)} = \frac{\omega_i^{(t+1)}}{\sum_{j=1}^{N} w_j^{(t+1)}}$$

3. Final Strong Classifier:
The strong classifier is a weighted majority vote of the $T$ weak classifiers:

$$H(x) = sign\left(\sum_{t=1}^{T} \alpha_t h_t(x)\right) \quad (5)$$

### 3.1.6. Logistic Regression

LR is a statistical method for modeling the probability of a binary outcome based on one or more predictor variables. It is widely used for classification problems where the dependent variable is categorical [59]

$$P(X) = \frac{e^{(b_0 + b_1^* X)}}{1 + e^{(b_0 + b_1^* X)}} \quad (6)$$

where $P(X)$ is the predicted output $b_0$ is the intercept term and $b_1$ is the coefficient of the single input value $x$. The logistic function transforms the linear combination of the predictors into a probability value between 0 and 1.

### 3.1.7. Multi-Layer Perceptron

MLP is a class of feedforward artificial neural networks. It consists of multiple layers of nodes (neurons) connected in a directed graph, where each layer is fully connected to the next one. MLPs are capable of learning complex patterns through supervised learning and are commonly used for tasks such as classification, regression, and function approximation [60].

Input Layer: Let $X = (x_1 + \cdots + x_n)$ be the input vector. Hidden Layer: For the $l^{th}$ hidden layer, let $h_i^{(l)}$ be the activation of $i^{th}$ neuron. The activation is computed as:

$$h_i^{(l)} = \varnothing^{(l)}\left(\sum_j \omega_{ij}^{(l)} x_j + b_i^{(l)}\right) \quad (7)$$

where $\varnothing^{(l)}$ is the activation function for the $l^{th}$ layer.

$\omega_{ij}^{(l)}$ is the weight connecting the $j^{th}$ neuron in the $(l-1)^{th}$ layer to the $i^{th}$ neuron in the $l^{th}$ layer.

$b_i^{(l)}$ is the bias term for the $i^{th}$ neuron in the $l^{th}$ layer.

$h_i^{(l-1)}$ is the activation of the $j^{th}$ neuron in the $(l-1)^{th}$ layer. For the input layer, $h_j^{(o)} = x_j$.

Output Layer: Let $y^k$ be the output of the $k^{th}$ neuron in the output layer. It is computed as:

$$y_k = \varnothing^{(L)}\left(\sum_j \omega_{kj}^{(L)} h_j^{(L-1)} + b_k^{(L)}\right) \quad (8)$$

We distinguish $\varnothing^{(1)}$ and $\varnothing^{(2)}$ because different layers may have different activation functions.

where:
- $L$ is the index of the output layer.
- $\phi(L)$ is the activation function for the output layer.

#### 3.1.8. Sentiment Analysis

Sentiment analysis is the process of classifying text into sentiment categories such as positive and negative [18,19]. There are two popular approaches to performing sentiment analysis, namely, the lexicon-based approach and the ML-based approach. The former approach requires the use of a corpus or dictionary to map words according to their semantic orientation into sentiment categories. The latter approach involves the use of labeled samples to train the ML algorithm for prediction. In this context, due to the unavailability of labeled datasets for training purposes, we consider the use of lexicon-based sentiment analysis suitable. We adopted Afinn developed in [61] based on usage and good performances shown across several sentiment analysis tasks. Afinn lexicon consists of 2477 English words, with each word associated with a sentiment score ranging from $-1$ to $-5$ for negative words and 1 to 5 for positive words. The sentiment score of a document is calculated by matching the Afinn lexicon wordlist to the document (at the word level). The associated sentiment score of the matched words is then aggregated to get the final sentiment score for the document. For illustration, if the final sentiment score is 1 or above, then the document sentiment score is positive, and it will be negative if the final sentiment score is $-1$ or below. Table 1 below presents the statistics of Afinn.

**Table 1.** Afinn statistics.

| Label | Count | Example |
| --- | --- | --- |
| Positive | 878 | Good |
| Negative | 1599 | Cry |

#### 3.2. Datasets

This study adopted two datasets to account for bias of algorithms towards a particular dataset. For identification purpose, we renamed the datasets as D1 and D2. Subsequent sections will provide a summary of the datasets.

#### 3.2.1. Dataset 1 (D1)

The dataset D1 was crawled from Reddit [62] and was annotated by two domain experts with labels, namely, not depressed, moderate, and severe. The label "not depressed" indicates the text shows no sign of depression, while "moderate" and "severe" indicate this is a depressive text, and the strength of depression in the text with "severe" being high. The dataset has been used in [63–65]. Specifically, it was used in the competition of the Second Workshop on Language Technology for Equality, Diversity, and Inclusion [66]. Authors in [66] pre-processed the data by removing the duplicates to create the final dataset (10,251 posts). The dataset was divided into three partitions, namely, train (6006 posts), dev (1000 posts), and test (3245 posts). The evaluation of model performances was performed using the dev dataset (as utilized by previous studies) to enable fair comparison with prior research work. The dataset is publicly available and can be accessed via this link: https://github.com/rafalposwiata/depression-detection-lt-edi-2022/tree/main/data (accessed on 12 November 2023). Table 2 below shows the summary of the data.

**Table 2.** D1 statistics.

| Label | Count$_{raw}$ | Count$_{preprocessed}$ | Example |
| --- | --- | --- | --- |
| Not depressed | 4649 | 3503 | Happy New Years Everyone: We made it another year |
| Moderate | 10,494 | 5780 | My life gets worse every year. That's what it feels like anyway |
| Severe | 1489 | 968 | Words can't describe how bad I feel right now: I just want to fall asleep forever. |

#### 3.2.2. Datasets 2 (D2)

The dataset, D2, was prepared by the authors in [37]. In their study, they reconstructed the dataset of [67] obtained from Reddit into four depression severity levels covering the clinical depression standards on social media. The dataset (3553 posts) was labeled manually by two human annotators, and an additional annotator was employed for cases with label disagreement. The dataset was divided into two partitions, namely, train (70%) and test (30%). The evaluation of model performances was performed using the test set. Table 3 below shows the data statistics.

**Table 3.** D2 statistics.

| Label | Count$_{raw}$ | Example |
|---|---|---|
| Minimal | 2587 | I just got out of a four year, mostly on but sometimes off relationship. The last interaction we had; he was moving out. The night before, he had strangled me. We've had a toxic relationship, but mostly loving. He truly tried to love me as much as possible but would get drunk and be verbally abusive. |
| Mild | 290 | I just feel like the street life has fucked my head up. There's so much I don't even know how to talk about anymore, I just hold that shit. The only person I can really chat with is a pal I know at the bar. He has PTSD and shit from the military bad, hard-up alcoholic nowadays after killing people. We talk once every few weeks and we are open and it's cool. But normal people? |
| Moderate | 394 | Sometimes, when I finally got out of bed and stood up, I felt like "Ugh, *finally*". Still, it did not happen every morning, and even when it did, I still felt rested from the long sleep, so I thought no more of it. Also, they were never nightmares. Sadly, my body got habituated to the sleep-component of Mirtazapine after about five months, and my old, warped sleep cycle slowly creeped back into my life. The only benefit left in the medicine was the mild mental cushioning it provided, but at the same time I started to suspect that what I needed wasn't cushioning but to make new constructive life decisions, that only I could make. |
| Severe | 282 | I know that I can't be unemployed forever but I'm just too anxious to really do anything. And everyone in my family keeps asking what my plan is and I keep lying because saying I've got nothing is just too humiliating. I'm just stuck. Have any of you have gone through something similar, and have any advice? I appreciate it. |

## 4. Results and Discussion

This section presents the results of our experiments. The evaluation of the algorithms in terms of performance is reported using metrics, namely, accuracy (A), weighted precision (P), weighted recall (R), and weighted F1 score (F). It is worth stating that due to the presence of class imbalance issues in datasets D1 and D2, this study emphasizes the P, R, and F metrics for evaluation more. Table 4 presents the comparative experimental results of the traditional ML algorithms fitted using the TF-IDF numerical vectors.

**Table 4.** The evaluation result of ML algorithms.

| | D1 | | | | D2 | | | |
|---|---|---|---|---|---|---|---|---|
| Algorithms | A | P | R | F | A | P | R | F |
| LR (TF-IDF) | 0.37 | 0.42 | 0.36 | **0.38** | **0.74** | **0.69** | **0.73** | **0.67** |
| NB (TF-IDF) | 0.36 | 0.40 | 0.36 | 0.36 | 0.72 | 0.52 | 0.71 | 0.60 |
| SVM (TF-IDF) | **0.43** | **0.50** | **0.43** | 0.31 | 0.72 | 0.56 | 0.71 | 0.60 |
| GBM (TF-IDF) | 0.39 | 0.46 | 0.39 | 0.35 | 0.73 | 0.67 | 0.72 | 0.66 |

In the D1 experimentation, results showed that SVM achieved the best outcome in terms of precision and recall. However, LR showed a better F1 score. It is worth noting that in terms of computational time, SVM took a very long time, GBM took a moderate time, and LR and NB were fast. Overall, the traditional ML algorithms performed poorly, most especially with the "severe" class. The "not depressed" class showed the best F1 score of all the three classes. This is unsurprising, considering that the "severe" class is the minority class. In the D2 experimentation, results showed that LR outperformed other models. In general, the traditional ML algorithms achieved good results. Across both datasets (D1 and D2), the LR showed an able-to-compete performance. However, the performance was generally poor in D1. Thus, our results showed that the traditional ML models are unable to cater to the diversity and class imbalance in the dataset. Furthermore, we performed the second experiment, in which we used the word embedding vectors (of BERT and SBERT) and adopted both the traditional ML algorithms (LR, SVM, and GBM) and deep learning algorithms (BiLSTM and BiGRU) as classifiers. The results of the second experiment are shown in Table 5 below.

**Table 5.** Evaluation result of the LLMs.

|                    | D1   |      |      |      | D2   |      |      |      |
|--------------------|------|------|------|------|------|------|------|------|
| Algorithms         | A    | P    | R    | F    | A    | P    | R    | F    |
| BERT + LR          | 0.63 | 0.63 | 0.61 | 0.62 | 0.72 | 0.67 | 0.72 | 0.69 |
| BERT + SVM         | 0.65 | 0.66 | 0.64 | 0.63 | 0.72 | 0.59 | 0.72 | 0.61 |
| BERT + GBM         | 0.65 | 0.67 | 0.65 | 0.63 | 0.72 | 0.64 | 0.72 | 0.67 |
| BERT + BiGRU       | 0.61 | 0.67 | 0.61 | 0.58 | 0.69 | 0.68 | 0.69 | 0.68 |
| BERT + BiLSTM      | 0.61 | 0.68 | 0.61 | 0.60 | 0.69 | **0.69** | 0.69 | 0.69 |
| BERT + Ensemble    | 0.66 | 0.68 | **0.66** | 0.64 | 0.73 | 0.66 | 0.73 | 0.68 |
| SBERT + LR         | 0.64 | 0.65 | 0.64 | 0.63 | 0.74 | **0.69** | 0.74 | 0.69 |
| SBERT + SVM        | 0.65 | 0.66 | 0.65 | 0.63 | 0.74 | 0.68 | 0.74 | 0.66 |
| SBERT + GBM        | 0.65 | 0.64 | 0.63 | 0.62 | 0.73 | 0.65 | 0.73 | 0.66 |
| SBERT + BiGRU      | 0.61 | 0.63 | 0.61 | 0.62 | 0.71 | **0.69** | 0.72 | **0.70** |
| SBERT + BiLSTM     | 0.61 | 0.62 | 0.61 | 0.61 | 0.73 | **0.69** | 0.74 | **0.70** |
| SBERT + Ensemble   | **0.69** | **0.69** | 0.65 | **0.68** | **0.76** | **0.69** | **0.75** | **0.70** |

In the D1 experimentation, results show that the use of word embedding vectors yields an improvement across all evaluation metrics in D1. However, the models show marginal improvement over each other. Furthermore, our approach performed best in terms of precision and F1 score. In the D2 experimentation, our approach showed improvements in terms of recall and F1 score. However, it is worth stating that improvements in the model performance were marginal when compared to the first experimental results presented in Table 4. This can be attributed to the proportionate distribution of the classes except the "minimal" class. Across the datasets (D1 and D2), our approach produced the best performance. This evidences the consistency, suitability, and efficiency of our approach.

Results from Table 6 show that incorporating SA as a feature enhanced the performance of our proposed model. Specifically, the results from the evaluation metrics show that the performances of our proposed model were improved with at least 0.02 (precision) and 0.01 (F1 score) in D1 and at least 0.08 (precision) and 0.06 (F1 score) in D2 in comparison to other models without SA feature. This improvement is consistent with the findings of Muñoz and Iglesias [63]. It is worth stating that models with the word embeddings (BERT and SBERT) showed better performance across D1 and D2 than models with TF-IDF. This is due to the ability of the transformer-based word embedding models to capture the relationship and contextual representation (language understanding) of the features. Overall, our approach performed best in D1 in terms of recall and F1 score. However, showed the best results across all metrics in D2. This further evidences the robustness, consistency, and efficiency of our approach. In comparison to other studies, our approach shows better performance than existing results, as shown in Tables 7 and 8 below.

**Table 6.** Evaluation results using LLMs (with sentiment analysis).

|                              | D1   |      |      |      | D2   |      |      |      |
|------------------------------|------|------|------|------|------|------|------|------|
| Algorithms                   | A    | P    | R    | F    | A    | P    | R    | F    |
| BERT + LR$_{AFINN}$          | 0.63 | 0.64 | 0.63 | 0.63 | 0.66 | 0.71 | 0.68 | 0.70 |
| BERT + SVM$_{AFINN}$         | 0.66 | **0.72** | 0.66 | 0.62 | 0.73 | 0.67 | 0.72 | 0.66 |
| BERT + GBM$_{AFINN}$         | 0.65 | 0.67 | 0.66 | 0.63 | 0.72 | 0.66 | 0.72 | 0.67 |
| BERT + BiGRU$_{AFINN}$       | 0.65 | 0.68 | 0.64 | 0.61 | 0.69 | 0.66 | 0.67 | 0.68 |
| BERT + BiLSTM$_{AFINN}$      | 0.65 | 0.67 | 0.64 | 0.65 | 0.72 | 0.70 | 0.73 | 0.71 |

**Table 6.** *Cont.*

| Algorithms | D1 | | | | D2 | | | |
|---|---|---|---|---|---|---|---|---|
| | A | P | R | F | A | P | R | F |
| BERT + Ensemble$_{AFINN}$ | 0.71 | 0.69 | 0.65 | 0.67 | 0.74 | 0.65 | 0.71 | 0.67 |
| SBERT + LR$_{AFINN}$ | 0.64 | 0.65 | 0.63 | 0.64 | 0.75 | 0.71 | **0.74** | 0.72 |
| SBERT + SVM$_{AFINN}$ | 0.65 | 0.65 | 0.65 | 0.63 | 0.74 | 0.72 | 0.70 | 0.66 |
| SBERT + GBM$_{AFINN}$ | 0.64 | 0.65 | 0.64 | 0.63 | 0.73 | 0.65 | 0.72 | 0.67 |
| SBERT + BiGRU$_{AFINN}$ | 0.60 | 0.61 | 0.58 | 0.60 | 0.71 | 0.66 | 0.70 | 0.68 |
| SBERT + BiLSTM$_{AFINN}$ | 0.60 | 0.61 | 0.58 | 0.59 | 0.73 | 0.70 | 0.72 | 0.71 |
| SBERT + Ensemble$_{AFINN}$ | **0.74** | 0.71 | **0.68** | **0.69** | **0.83** | **0.77** | **0.74** | **0.76** |
| XLNet + Ensemble$_{AFINN}$ | 0.67 | 0.70 | **0.68** | 0.66 | 0.75 | 0.67 | 0.72 | 0.71 |
| ALBERT + Ensemble$_{AFINN}$ | 0.67 | 0.68 | 0.66 | 0.64 | 0.72 | 0.64 | 0.71 | 0.70 |
| RoBERTa + Ensemble$_{AFINN}$ | 0.69 | 0.68 | 0.66 | 0.67 | 0.75 | 0.67 | 0.72 | 0.71 |

**Table 7.** Performance comparison to previous studies (D1).

| Literature | Method | A | P | R | F |
|---|---|---|---|---|---|
| Poswiata and Perelkiewicz [66] | Ensemble (averaging predictions of RoBERTa and DepRoBERTa) | 0.69 | 0.66 | 0.62 | 0.63 |
| Muñoz and Iglesias [63] | Ensemble (Lexicon + all-distilroberta-v1) | - | 0.58 | 0.61 | 0.58 |
| Our study | SBERT + Ensemble$_{AFINN}$ | **0.74** | **0.71** | **0.68** | **0.69** |

**Table 8.** Performance comparison to previous studies (D2).

| Literature | Method | A | P | R | F |
|---|---|---|---|---|---|
| Muñoz and Iglesias [63] | Ensemble (Lexicon + all-distilroberta-v1) | - | 0.75 | 0.73 | 0.74 |
| Ilias et al. [68] | Incorporated linguistic feature into MentalBERT | - | 0.73 | 0.73 | 0.73 |
| Our study | SBERT + Ensemble$_{AFINN}$ | **0.83** | **0.77** | **0.74** | **0.76** |

## 5. Conclusions

　　　This paper aimed to develop a depression detection model using a well-performing ML algorithm. To this end, we compared several ML algorithms ranging from the traditional ML to the deep learning algorithms. Furthermore, we utilized TF-IDF and word embeddings from the LLMs to generate the numeric vectors fitted into the ML models. Specifically, we compared the influence of sentiment indicators as an external feature in the models. Our experimental results evidenced that the use of SBERT and stacking ensemble model achieved SOTA results. Our result is beneficial to the medical/healthcare stakeholders and practitioners for early depression detection. This intervention helps in making a more accurate diagnosis and monitoring the symptoms over time.

　　　Theoretically, we have evidence that sentiment analysis indicator is an important feature for enhancing the performance of depression detection. Secondly, we have evidence that the word embedding models from the transformer architectures yield an ability to compete for performance across several datasets. Thus, it can be relied on as an off-the-shelf depression detection model. To conclude, we have presented the use of stacked ensemble ML models with SBERT and sentiment indicators for the prediction of depression, with improved metrics when compared to previous works. However, SBERT, as with any BERT model, can struggle to generate embeddings when sentences become too long. Also, it may struggle with context-aware tasks, where there might be subtle semantics differences in sentences. To this end, in future works, we will experiment with newer, larger language models such as GPT, LLaMa, and PaLM variants with larger trainable parameters, as recent research [69] has shown to have better sentiment scores when compared to BERT. Also, we will explore the explainability of the black box SBERT model. For example, we need to understand how the Siamese BERT models update their weight during training, having trained with different input sentences but using the same weight for each training epoch. Furthermore, it is worth stating that our experimental results relied on Reddit datasets; as such, we envisage evaluating our methodology with datasets from other social networks. Similarly, our future work will consider incorporating emotion recognition features into our methodology. Lastly, we recommend extending our methodology to the detection of harmful content online and the detection of academic stress.

**Author Contributions:** Conceptualization, B.O.; methodology, B.O., H.S. and O.S.; software, B.O.; validation, B.O. and O.S.; formal analysis, B.O.; investigation, B.O.; resources, B.O.; data curation, B.O.; writing—original draft preparation, B.O., H.S. and O.S.; writing—review and editing, B.O., H.S. and O.S.; visualization, B.O., H.S. and O.S.; supervision, B.O.; project administration, B.O. and O.S. All authors have read and agreed to the published version of the manuscript.

**Funding:** This research received no external funding.

**Institutional Review Board Statement:** Not applicable.

**Informed Consent Statement:** Not applicable.

**Data Availability Statement:** All the data are presented in the study.

**Conflicts of Interest:** The authors declare no conflicts of interest.